\RequirePackage{fix-cm}

\documentclass[twocolumn]{svjour3}

\smartqed
  
\usepackage{amsmath}
\usepackage{hyperref}
\usepackage{cite}
\usepackage{doi}
\usepackage{multirow}
\usepackage{algorithm}
\usepackage{algpseudocode}
\usepackage{arydshln}
\usepackage{afterpage}
\usepackage{subcaption}
\usepackage{graphicx}
\usepackage{color}
\usepackage{epstopdf}
\usepackage{bm}
\usepackage{textcomp}
\usepackage{microtype}
\usepackage{tikz}
\usepackage[toc,page]{appendix}
\usepackage[T1]{fontenc}

\algnewcommand{\IIf}[1]{\State\algorithmicif\ #1\ \algorithmicthen}
\algnewcommand{\ElseIIf}[1]{\State\algorithmicelse\ #1}
\algnewcommand{\EndIIf}{\unskip\ \algorithmicend\ \algorithmicif}
\DeclareMathOperator*{\argmin}{arg\,min}
\algrenewcommand\alglinenumber[1]{\scriptsize #1:}
\def\NoNumber#1{{\def\alglinenumber##1{}\State #1}\addtocounter{ALG@line}{-1}}
\algnewcommand{\LineComment}[1]{\hspace{1cm} \NoNumber{\(\triangleright\) #1}}

\begin{document}
\title{Two-phase Optimization of Binary Sequences\\with Low Peak Sidelobe Level Value}
\titlerunning{Two-phase Optimization of Binary Sequences with Low PSL Value}
\author{Borko Bo\v{s}kovi\'{c} \and Janez Brest}
\institute{B. Bo\v{s}kovi\'{c} \and J. Brest \at
    Faculty of Electrical Engineering and Computer Science,\\ University of Maribor, SI-2000 Maribor, Slovenia \\
    \email{borko.boskovic@um.si, janez.brest@um.si}
}
\renewcommand{\labelitemi}{$\bullet$}

\date{\today}

\maketitle

\begin{abstract}
The search for binary sequences with low peak sidelobe level value represents a formidable computational problem.
To locate better sequences for this problem, we designed a stochastic algorithm that uses two fitness functions.
In these fitness functions, the value of the autocorrelation function has a different impact on the final fitness
value. It is defined with the value of the exponent over the autocorrelation function values. Each function is used
in the corresponding optimization phase, and the optimization process switches between these two phases until the stopping 
condition is satisfied. The proposed algorithm was implemented using the compute unified device architecture and 
therefore allowed us to exploit the computational power of graphics processing units. This algorithm was tested on
sequences with lengths $L = 2^m - 1$, for $14 \le m \le 20$. From the obtained results it is evident that the usage of
two fitness functions improved the efficiency of the algorithm significantly, new-best known solutions were achieved,
and the achieved PSL values were significantly less than~$\sqrt{L}$.
\end{abstract}

\section{Introduction}
\label{sec:intro}
The binary sequences with low peak sidelobe level value have many applications in diverse areas such as wireless
communication, cryptography, and radar applications \cite{golomb05,kroszczynski69}. The construction and computational
approaches are used for solving this problem \cite{Song15}. In this paper, we present a computational approach that
uses a stochastic algorithm. In contrast to exhaustive search, our approach cannot provide optimal solutions but
in a reasonable time we can locate optimal or near-optimal solutions. Therefore, our approach is also suitable for
solving larger instances of the problem. The binary sequence of length $L$ in our problem is defined as follows:

\begin{equation}
\begin{split}
S_L &=\{s_1, s_2, ..., s_L\}\\
s_i &\in \{+1, -1\}~;~~~
i = \{1, 2, ..., L\}.
\label{eq:seq}
\end{split}
\end{equation}

\noindent
The autocorrelation function at shift $k$ of binary sequence $S_L$ is shown in Eq.~(\ref{eq:ck}) while the peak sidelobe level (PSL)
is shown in Eq.~(\ref{eq:psl}).

\begin{equation}
 C_k(S_L)=\sum_{i=1}^{L-k}s_i\cdot s_{i+k}~;~~~ k \in \{0, 1, ..., L-1\}
 \label{eq:ck}
\end{equation}

\begin{equation}
 \mathit{PSL}(S_L) = \max\limits_{1<k<L}|C_k(S_L)|
 \label{eq:psl}
\end{equation}

\noindent
The main goal of a binary sequences problem with low peak sidelobe level is to find an optimal sequence that
has the minimal PSL value, as shown in Eq.~(\ref{eq:goal}).

\begin{equation}
 S^{*}_L = \argmin\limits_{S_L\in B_L} \mathit{PSL}(S_L)
 \label{eq:goal}
\end{equation}

\noindent
In this equation, $S^{*}_L$ stands for the binary sequence with the optimal value of PSL and $B_L$ is a set of all
sequences with length $L$. From Eq.~(\ref{eq:seq}) it is evident that the number of sequences with length $L$ is $2^L$. With
exhaustive search, it is possible to locate optimal sequences for small instances of the problem. For example, an exhaustive
search has been applied to $L = 64$~\cite{Coxson05}. The exhaustive search was also applied under the restriction of 
$m$-sequence~\cite{Dimitrov21a}. However, stochastic algorithms are useful for longer sequences~\cite{Dimitrov20a,Brest21}.
To obtain good results, stochastic algorithms need a fitness function that guides the search process throughout the search space.
Different fitness functions have already been used for different values of $L$ in recent works \cite{Dimitrov21a,Brest21}. According
to this observation, we propose an algorithm that uses two fitness functions within the optimization process. These functions are
the same, but with different values of exponent $\alpha$, as shown in Eq.~(\ref{eq:f}).

\begin{equation}
 \mathit{F}(S_L) = \sum_{k=1}^{L-1}|C_k(S_L)|^\alpha
 \label{eq:f}
\end{equation}

\noindent
The value of exponent $\alpha$ determines which values of autocorrelation function $C_k(S_L)$ are more important within 
the optimization process. In the first fitness function, we use a smaller value of $\alpha$, and both functions are used in
the corresponding optimization phase. Each function guides the search process differently, and, consequently, we can obtain
better results. Therefore the optimization process switches between these two phases until the stopping condition is
satisfied. The proposed algorithm has been implemented using Compute Unified Device Architecture (CUDA), and therefore
allows us to exploit the computational power of Graphics Processing Units (GPU). In such a way, the solver speed is
increased significantly and better results can be obtained at the same time. Although our solver is suitable for solving
any larger instance of the problem, it was tested on sequences with lengths $L = 2^m - 1$, for $14 \le m \le 20$. These
sequences are used frequently in the literature which enables us to compare our results with other approaches. From the obtained
results it is evident that the usage of two fitness functions and GPU improves the efficiency of the solver significantly,
new-best known solutions were obtained for all selected instances of the problem, and the achieved PSL values were significantly
less than $\sqrt{L}$. As the instance size increased, the difference between the old best-known and achieved PSL
values increased. Based on the described observations, the main contributions of this paper are:

\begin{itemize}
 \item The two-phase optimization of binary sequences with low peak sidelobe level value 
  that guides the search process according to the two fitness functions.
 \item The concurrent computing on the graphics processing unit enables a significant improvement of the solver speed
 or the number of sequence evaluations per second.
 \item The new best-known sequences achieved by the proposed algorithm for all selected instance sizes.
\end{itemize}

The remainder of the paper is organized as follows. Related work is described in Section~\ref{sec:related}.
The proposed two-phase optimization algorithm is given in Section~\ref{sec:optimization}. The description of the 
experiments, analysis of the proposed algorithm, and the obtained results are presented in Section~\ref{sec:experiment}.
Finally, the paper ends with a conclusion in Section~\ref{sec:conclusion}.

\section{Related work}
\label{sec:related}
Over the years different approaches have been used successfully for the problem of binary sequence with low PSL value. 
Computational approaches use exhaustive search and stochastic algorithms. Exhaustive search explorers the
entire search space systematically, and therefore it can provide the optimal solution. An efficient exhaustive search was implemented
in \cite{Leukhin13} and optimal binary sequences with minimum PSL values were achieved for instances up to $L=72$. An exhaustive
search with restriction of $m$-sequences was used in \cite{Dimitrov21a}. Within this approach, all PSL-optimal
Legendre sequences were revealed with or without rotations. The obtained PSL values of all revealed Legendre sequences
were strictly greater than $\sqrt{L}$, for $235723 < L < 432100$.

In contrast to exhaustive search, stochastic neighborhood search is also useful for many problems \cite{Vincent21,Longqing21,Biao21},
including locating sequences with low PSL value. Recent
approaches~\cite{Dimitrov20a}, \cite{Dimitrov20b}, and~\cite{Brest21} were able to reach sequences that have new best-known
PSL values, for $ 105 < L \leq 10^6$. The fitness function (see Eq.~\ref{eq:f}) was used in these approaches to
guide the search process throughout the search space, and different fitness functions had already been used for different values of $L$.
The stochastic approach was also used in~\cite{Gallardo09,Boskovic17,Brestarxiv,Dimitrov21b} for low-autocorrelation binary sequences with
merit factor. The self-avoiding walk and memetic algorithm were used in these works and new-best known sequences were also reported.
In contrast to our problem, the main goal of the low-autocorrelation binary sequences problem with merit factor is to locate sequences
with the maximum merit factor.  It is defined as shown by the following equation:
\begin{equation}
\mathit{MF}(S) = \frac{L^2}{2 \cdot \sum_{k=1}^{L-1}C_k^2(S)}\nonumber
\end{equation}

Unlike the described approaches, our approach used two-phase optimization. In our previous work~\cite{Boskovic20}, a two-phase optimization process
was applied to the protein folding problem on a three-dimensional AB off-lattice model. The fitness function that was used in the first
optimization phase was responsible for forming conformations with a good hydrophobic core. The fitness function that was used in the
second phase was responsible for final optimization. Similarly, in this work, we used two fitness functions that use different values of
exponent $\alpha$ (see Eq.~\ref{eq:f}). With this exponent, we determined which values of autocorrelation function are more important within
the optimization phases.

To get an even more efficient solver, we used the computational power of the GPU. The author in~\cite{Dimitrov21a} also used a GPU to exhaust
the search space of all $m$-sequences with $L = 2^m - 1$, for $18 \le m \le 20$. In~\cite{Dominik21}, the authors used the parallel nature of the
GPU, and provided an effective method of solving the the low-autocorrelation binary sequences problem with merit factor. From these works it is
evident that the GPU allows development of computationally efficient solvers.

\section{Two-phase optimization}
\label{sec:optimization}
\begin{algorithm}[t]
    \scriptsize
    \caption{Search algorithm}
    \label{alg:search}
    \begin{algorithmic}[1]
    \Procedure{Search}{$L,\mathit{SEED}, \mathit{FLIP}_{\mathit{lmt}}, \mathit{LS}_{\mathit{lmt}}, N_{\mathit{lmt}}, \alpha_1, \alpha_2$}
    \State $pivot \gets $ \Call{Rand}{$L,\mathit{SEED}$}
    \State $f \gets \sum_{k=1}^{L-1}|C_k(S_L)|^{\alpha}$ \label{alg:a1}
    \State $\alpha \gets \alpha_1$ \label{alg:a2}
    \LineComment{Concurrent computing}
    \State $\mathit{value}_{\mathit{local}} \gets $ \Call{Fitness}{$pivot,f,\alpha$} \label{alg:p1}
    \While{not-stopping condition}
        \State $\mathit{start} \gets$ \Call{Rand}{~} \% $L$
        \LineComment{Concurrent computing}
        \State \{$\mathit{value},\mathit{psl}$\} $\gets$ \Call{Neighborhood}{$\mathit{pivot}, f, \alpha, N_{\mathit{lmt}}, \mathit{start}$} \label{alg:p2}
        \State $\mathit{value}_{\mathit{step}} \gets $MAX\_VALUE
        \For{$i \gets 1,\, N_{\mathit{lmt}}$}
            \State $j \gets (\mathit{start}+i) \% L$
            \If {$\mathit{value}_j < v\mathit{alue}_{\mathit{step}}$}
                \State $\mathit{value}_{\mathit{step}} \gets \mathit{value}_{j}$
                \State $k \gets j$
            \EndIf
            \If {$\mathit{psl}_{j} < \mathit{psl}_{\mathit{best}}$}
                \State $\mathit{psl}_{\mathit{best}} \gets \mathit{psl}_{j}$
                \State $\mathit{solution}_{\mathit{best}} \gets \mathit{neighbour}_{j}$ \label{alg:bs}
            \EndIf
        \EndFor
        \State $\mathit{pivot} \gets $ \Call{Flip}{$\mathit{pivot},k$}
        \If {$\mathit{value}_{\mathit{step}} < \mathit{value}_{\mathit{local}}$}
            \State $\mathit{unimproved} \gets 0$
            \State $\mathit{local}_{\mathit{best}} \gets \mathit{pivot}$
            \State $\mathit{value}_{\mathit{local}} \gets \mathit{value}_{\mathit{step}}$
        \Else
            \If {$\mathit{value}_{\mathit{step}} > \mathit{value}_{\mathit{local}}$}
                \State $\mathit{unimproved} \gets \mathit{unimproved} + 1$
                \LineComment{Switch the optimization phase}
                \If {$\mathit{unimproved} > \mathit{LS}_{\mathit{lmt}}$} \label{alg:sw}
                    \State $\mathit{pivot} \gets $\Call{Rand\_flip}{$\mathit{local}_{\mathit{best}},\mathit{FLIP_{\mathit{lmt}}}$}
                    \IIf {$\alpha = \alpha_1$} {$\alpha \gets \alpha_2$} 
                    \ElseIIf {$\alpha \gets \alpha_1$} \EndIIf
                    \LineComment{Concurrent computing}
                    \State $\mathit{value}_{\mathit{local}} \gets $ \Call{Fitness}{$\mathit{pivot},f,\alpha$} \label{alg:p3}
                    \State $\mathit{unimproved} \gets 0$
                \EndIf
            \EndIf
        \EndIf
    \EndWhile
    \EndProcedure
    \end{algorithmic}
\end{algorithm}

\begin{figure}[t]
    \centerline{\resizebox{0.49\textwidth}{!}{\input{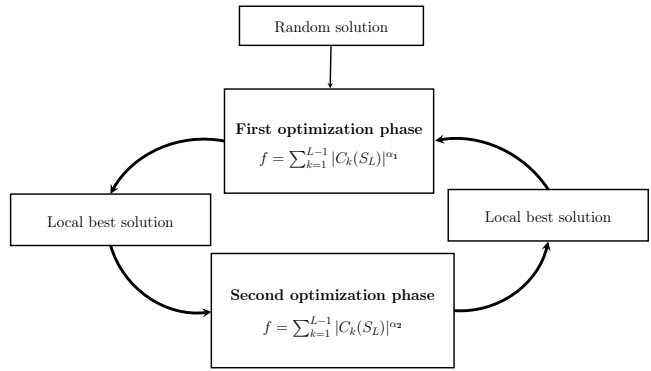}}}
    \caption{Two-phase optimization process of the proposed algorithm.}
    \label{fig:phases}
\end{figure}

The proposed algorithm contains two optimization phases, and each of them uses its fitness function, as shown in
Algorithm~\ref{alg:search}. The optimization process starts with the random pivot ($S_L$) and the first optimization phase.
The fitness function with exponent $\alpha_1$ is used (see lines~\ref{alg:a1} and \ref{alg:a2}) in this phase to
guide the optimization process. In every iteration of the {\it while} loop, a limited neighborhood search is performed, and
the best neighbor is selected as a new pivot. The limited neighborhood is determined with the randomly selected first 
neighbor and $N_{\mathit{lmt}}$ consecutive neighbors. Each neighbor differs from the pivot in only one element of
the sequence ($s_i$). The PSL values are also calculated for all selected neighbors, and the neighbor with the best
PSL value is saved as the best solution (see line~\ref{alg:bs}). Note that the search process is not guided 
according to the PSL values, but according to the values of the fitness function. If the local best solution is not improved
for $\mathit{LS}_{\mathit{lmt}}$ limited neighborhood searches, then the optimization process switches to the second 
optimization phase (see line~\ref{alg:sw}). In the process of phase switching, the fitness function is changed,
the local best individual of the previous phase is selected as the new pivot and its $\mathit{FLIP}_{\mathit{lmt}}$
elements are selected randomly and flipped. The flipped elements ensure that the algorithm is not trapped into
local optima. The described optimization process continues to switch between two phases until the stopping condition is
met, as shown in Fig.~\ref{fig:phases}. The different exponent values of fitness function determine which values of the
autocorelation function have more influence on the final fitness value, and, consequently, the path of the search process
at each optimization phase.

An efficient implementation of the fitness function and limited neighborhood search is crucial for our algorithm. For 
this purpose, we used a one element flip mechanism~\cite{Dimitrov20b,Gallardo09} and massively parallel computing with CUDA.
The parallel version of the fitness function (see lines \ref{alg:p1}, \ref{alg:p3}) calculates each component of the sum 
in Eq.~(\ref{eq:f}) concurrently, while the parallel version of limited neighborhood search (see line \ref{alg:p2}) 
calculates the whole Eq.~(\ref{eq:f}) concurrently for several neighbors.

\begin{table*}[ht]
 \centering
 \scriptsize
 \caption{The analyses of control parameters $\mathit{LS}_{\mathit{lmt}}$ and $\mathit{FLIP_{\mathit{lmt}}}$ with 30 independent runs on
 the GPU and with the following settings: $L=65535$, $\alpha_1=4$, $\alpha_2=13$, $N_{\mathit{lmt}}=6912$. The stopping condition was the
 number of sequence evaluations
 $\mathit{NSEs}=10^{10}$. The results in the Table represent the mean value of the achieved $PSL$ values.}
 \label{tab:par}
 \resizebox{0.65\textwidth}{!}{
 \begin{tabular}{r|rrrrrr}
   \multirow{2}{*}{$\mathit{FLIP_{\mathit{lmt}}}$} & \multicolumn{5}{c}{$\mathit{LS}_{\mathit{lmt}}$}\\
    & 4000 & {\bf 2000} & 1000 & 500 & 250 \\
   \hline
  20 & 212.767 & 212.867 & 213.067 & 212.933 & 213.333 \\
  {\bf 10} & 212.967 & {\bf 212.700} & 212.967 & 212.933 & 213.533\\
   5 & 212.967 & 212.967 & 213.067 & 213.100 & 213.233 \\
   2 & 212.800 & 212.967 & 212.833 & 213.000 & 213.233\\
   1 & 212.933 & 212.933 & 213.033 & 213.100 & 213.333\\
 \end{tabular}}
\end{table*}

\section{Experiments}
\label{sec:experiment}
The proposed algorithm was implemented for Central Processing Unit (CPU) and GPU by using 
Stochastic Problem-Solving Environment\footnote{Available at http://spse.feri.um.si/}.
This environment allows the rapid development and testing of stochastic algorithms for different problems in an efficient way.
Solvers were compiled with a GNU C++ compiler 9.3.0 and CUDA Toolkit 10.1. The CPU solver was running on an AMD Ryzen 5 3600 CPU with 16 GB RAM
under Ubuntu 20.04. The GPU solver was running on an NVIDIA A100 SXM4 within a grid environment VEGA\footnote{Available at http://www.sling.si/sling/}.
To evaluate the efficiency of the proposed algorithm, we used sequences with lengths $L = 2^m - 1$, for $14 \le m \le 20$. The size of these sequences 
was relatively large, and they were also used frequently in the literature. This enabled us to compare our results with other approaches.

\subsection{Control parameters}
In this section, we will analyze the influence of the control parameters on the algorithm’s efficiency. The proposed approach has the following 
control parameters:
\begin{itemize}
    \item $\mathit{LS}_{\mathit{lmt}}$ -- trigger of switching between optimization phases.
    \item $\mathit{FLIP_{\mathit{lmt}}}$ -- the number of randomly selected and flipped elements of pivot that will be used in the next optimization phase.
    \item $N_{\mathit{lmt}}$ -- the size of the neighborhood.
    \item $\alpha_1$ -- the fitness function exponent in the first optimization phase.
    \item $\alpha_2$ -- the fitness function exponent in the second optimization phase.
    \item $cuda$ -- use of CUDA architecture or concurrent computing on GPU.
\end{itemize}

\noindent
The neighborhood size ($N_{\mathit{lmt}}$) was selected according to the used CUDA architecture. The NVIDIA A100 SXM4 GPU 
architecture was used in our experiment. It had 6912 CUDA cores and this value was also used for the value of $N_{\mathit{lmt}}$.
This means the solver achieved
maximum computational efficiency because each neighbor is evaluated on one core simultaneously.

The values of control parameters $\mathit{LS}_{\mathit{lmt}}$ and $\mathit{FLIP_{\mathit{lmt}}}$ were analyzed on the GPU and with the following settings: $L=65535$,
$\alpha_1=4$, $\alpha_2=13$. Within this analyse, 30 independent runs were performed for each pair of $\mathit{LS}_{\mathit{lmt}}$ and 
$\mathit{FLIP_{\mathit{lmt}}}$ values, and each run was limited with the number of sequence evaluations $\mathit{NSEs}=10^{10}$. From the results shown
in Table~\ref{tab:par} we can observe that the best mean $PSL$ value (shown in bold-face) was obtained for $\mathit{LS}_{\mathit{lmt}}=2000$ and 
$\mathit{FLIP_{\mathit{lmt}}}=10$. These values were also used in all the remaining experiments.

\begin{table}[t]
\centering
\caption{The analyses of control parameters $\alpha_1$ and $\alpha_2$ with 30 independent runs on the GPU and with the following settings: $L=65535$,
$\mathit{LS}_{\mathit{lmt}}=2000$, $\mathit{FLIP_{\mathit{lmt}}}=10$, $N_{\mathit{lmt}}=6912$. The stopping condition was the number of sequence evaluations
 $\mathit{NSEs}=10^{10}$.}
\label{tab:alpha}
\resizebox{0.20\textwidth}{!}{
 \begin{tabular}{c|r}
  $\alpha_1$/$\alpha_2$ & $PSL_{mean}$ \\
  \hline
  {\bf 4/13} & {\bf 212.700} \\
  4/4 & 232.833 \\
  13/13 & 238.967 \\
 \end{tabular}}
\end{table}

\begin{table*}[t]
\centering
\caption{4 runs on GPU (NVIDIA A100 SXM4) that was limited with $\mathit{runtime} = 4$ days. The following settings were used: 
$\mathit{LS}_{\mathit{lmt}}=2000$, $\mathit{FLIP_{\mathit{lmt}}}=10$, $N_{\mathit{lmt}}=6912$, $\alpha_1=4$. The values of $\alpha_2$ are shown
in the Table.}
\label{tab:4days}
\resizebox{0.7\textwidth}{!}{
 \begin{tabular}{rrrr|rrrr}
  $m$ &  \multicolumn{1}{c}{$L$} & $\alpha_2$ & $\mathit{PSL}_{\mathit{old}}$ & $\mathit{PSL}_{\mathit{run_1}}$ & 
  $\mathit{PSL}_{\mathit{run_2}}$ & $\mathit{PSL}_{\mathit{run_3}}$ & $\mathit{PSL}_{\mathit{run_4}}$\\
  \hline
  14 &  16383 & 13 &       102\,\cite{Brest21} & {\bf 101} &       104 & {\bf 101} &      102 \\
  15 &  32767 & 13 &       149\,\cite{Brest21} &       152 & {\bf 146} &       149 &      149 \\
  16 &  65535 & 13 &       218\,\cite{Brest21} & {\bf 210} & {\bf 210} & {\bf 209} & {\bf 211} \\
  17 & 131071 & 13 &       323\,\cite{Brest21} & {\bf 301} & {\bf 302} & {\bf 303} & {\bf 302} \\
  18 & 262143 & 11 &   507\,\cite{Dimitrov21a} & {\bf 434} & {\bf 434} & {\bf 435} & {\bf 438} \\
  19 & 524287 & 10 &   731\,\cite{Dimitrov21a} & {\bf 628} & {\bf 629} & {\bf 628} & {\bf 629} \\
  20 & 1048575 & 10 & 1024\,\cite{Dimitrov21a} & {\bf 902} & {\bf 902} & {\bf 901} & {\bf 903} \\
 \end{tabular}}
\end{table*}

The control parameters $\alpha_1$ and $\alpha_2$ were used to demonstrate the efficiency of the two-phase optimization. With the same settings as in the
previous experiment, we obtained the results that are shown in Table~\ref{tab:alpha}. We can see that when the same fitness function was used 
throughout the entire optimization process (both parameters $\alpha_1$ and $\alpha_2$ had the same value) the obtained results were significantly worse in comparison
with the results of the two-phase optimization.

\begin{figure}[t!]
    \centering
    \includegraphics[width=0.40\textwidth]{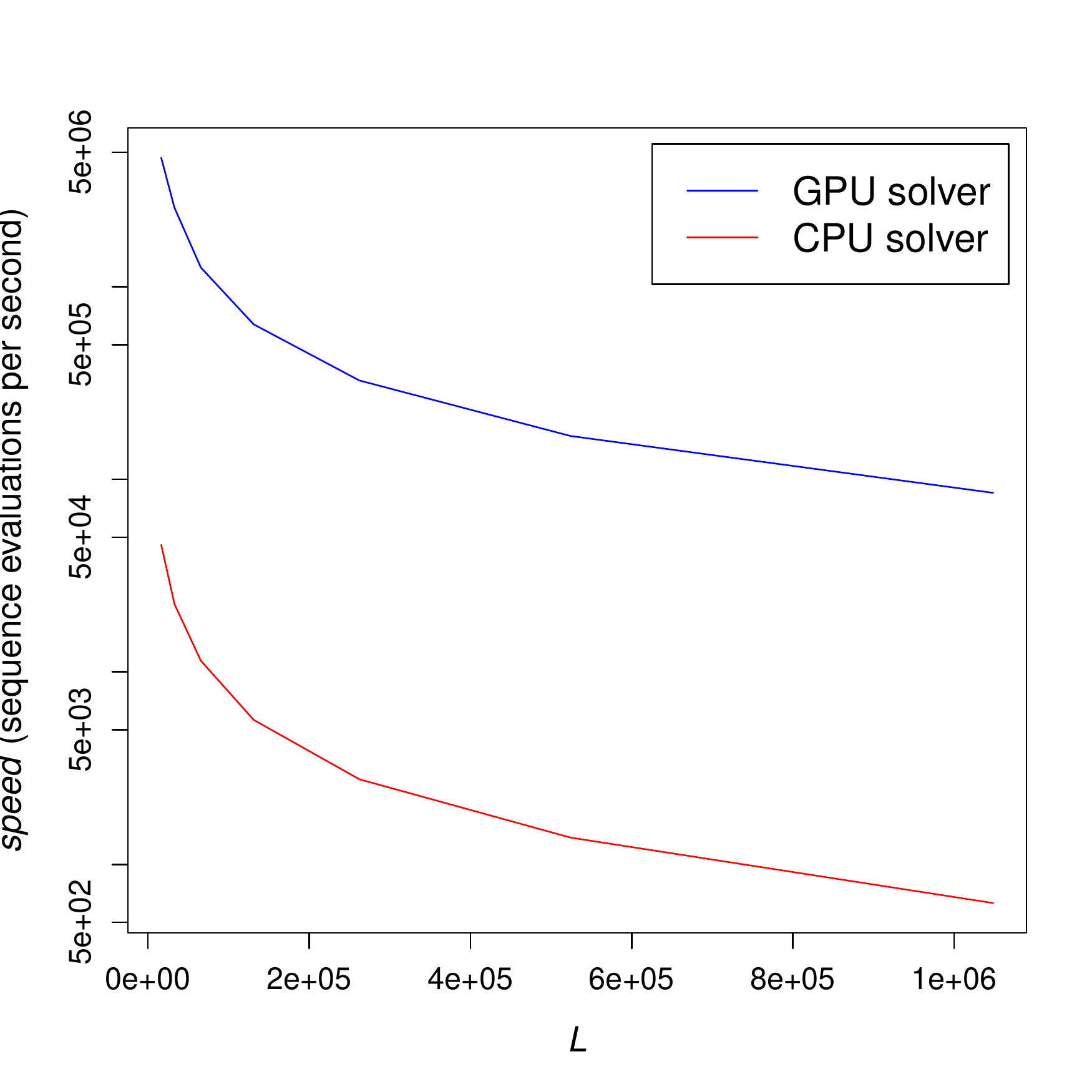}
    \caption{Speed of the CPU and GPU solvers (see Table~\ref{tab:speed}).}
    \label{fig:speed}
\end{figure}

To demonstrate the efficiency of concurrent computing on CUDA architecture, we implemented a CPU and a GPU solver. The parameter $cude$ determined which
solver would be selected. The solvers were different only in the concurrent implementation of lines~\ref{alg:p1}, \ref{alg:p2}, and \ref{alg:p3} in Algorithm~\ref{alg:search}.
Both solvers were analyzed on all selected instance sizes with the same settings. The CPU solver was run on a single AMD Ryzen 5 3600 CPU core, while the GPU solver
was run on 6912 NVIDIA A100 SXM4 CUDA cores. The speed or the number of sequence evaluations per second of both solvers is shown in Table~\ref{tab:speed} and Fig.~\ref{fig:speed}. As we
can see, the speedup of the GPU solver was more than 100. We can also see that the lowest speed of the GPU solver (for $L=1048575$) was better than the best speed
of the CPU solver (for $L=16383$). This indicates that the usage of CUDA architecture allows significant speedup and, therefore, we can reach better results at the
same time.

\begin{table}[t!]
    \centering
    \caption{Speed of the CPU and GPU solvers. One run was performed for each $L$ by each solver with the following settings:
    $\mathit{LS}_{\mathit{lmt}}=2000$, $\mathit{FLIP_{\mathit{lmt}}}=10$, $N_{\mathit{lmt}}=6912$, $\alpha_1=4$, $\alpha_2=10$, $\mathit{runtime}_{\mathit{lmt}}=1h$.}
    \label{tab:speed}
    \resizebox{0.49\textwidth}{!}{
    \begin{tabular}{r|rrr}
    $L$ & CPU solver & GPU solver & speedup\\
    \hline
    16383 & 45705 & 4678451 & 102.36 \\
    32767 & 22530 & 2577101 & 114.39 \\
    65535 & 11425 & 1258873 & 110.18 \\
    131071 & 5632 & 638043 & 113.29 \\
    262143 & 2769 & 326148 & 117.78 \\
    524287 & 1378 & 167861 & 121.81 \\
    1048575 & 640 & 85062 & 132.91 \\
    \end{tabular}}
\end{table}

\subsection{The new best-known PSL values}
To demonstrate the superiority of our algorithm in comparison with other algorithms, the best PSL values were compared for all selected instance sizes.
For this purpose, we performed 4 runs for each instance size on the GPU. All runs were limited with $\mathit{runtime} = 4$ days, and with the following
settings: $\mathit{LS}_{\mathit{lmt}}=2000$,
$\mathit{FLIP_{\mathit{lmt}}}=10$, $N_{\mathit{lmt}}=6912$, $\alpha_1=4$. The values of $\alpha_2$ were set according to the value of $L$. For the two largest 
instances it was set to 10, for the third-largest instance it was set to 11, and, for all others, it was set to 13. With these values, we prevented a floating-point overflow error
of fitness function values in the second optimization phase. The results of the described runs are shown in Table~\ref{tab:4days}.
It can be observed that new best-known PSL values were achieved for all instances, and as the instance size increased, the 
difference between the old best-known and achieved PSL values increased.

\begin{figure*}[h!]
     \centering
     \begin{subfigure}{0.49\textwidth}
         \centering
         \includegraphics[width=\textwidth]{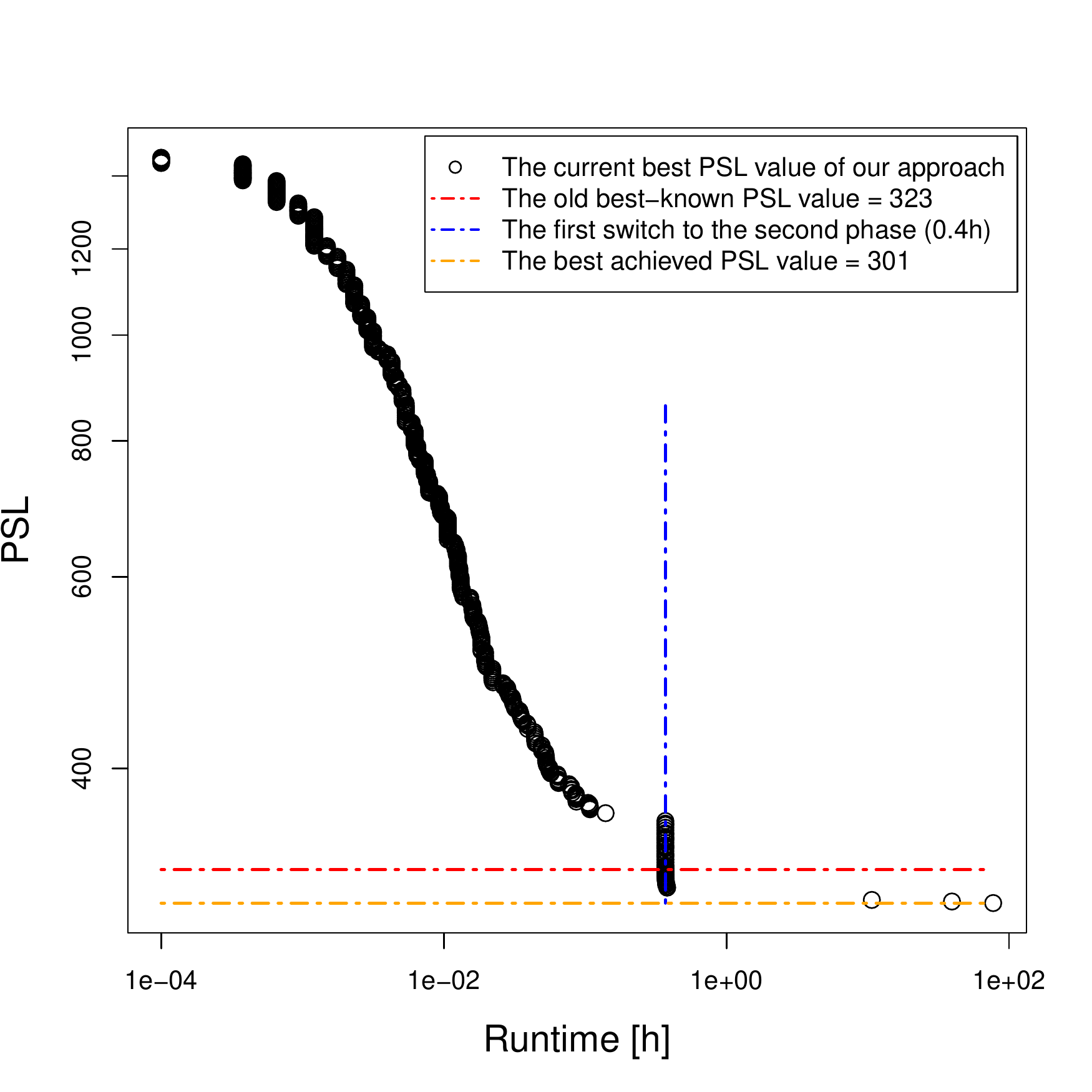}
         \caption{$L = 131071$}
         \label{fig:L131071}
     \end{subfigure}
     \hfill
     \begin{subfigure}{0.49\textwidth}
         \centering
         \includegraphics[width=\textwidth]{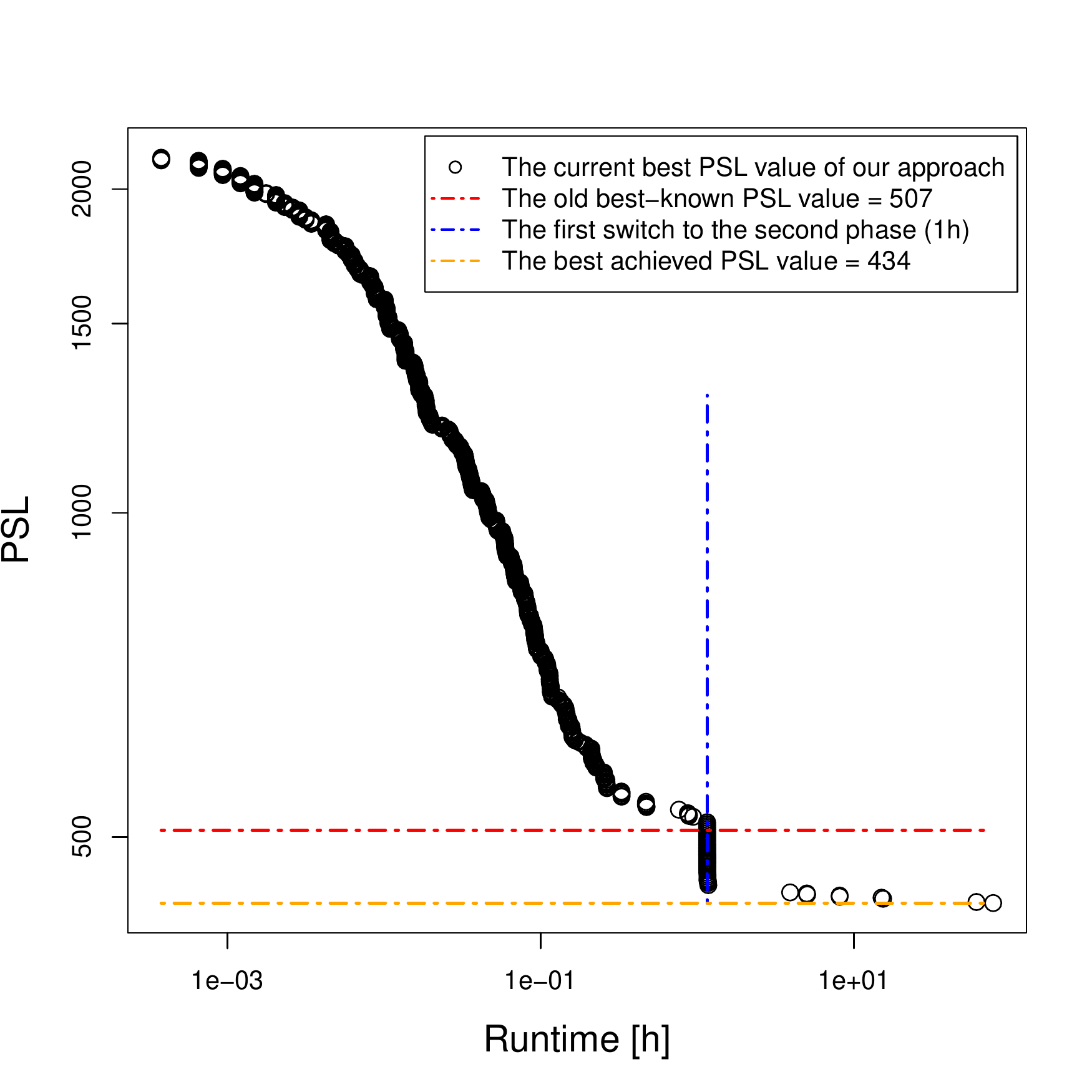}
         \caption{$L = 262143$}
         \label{fig:L262143}
     \end{subfigure}

     \begin{subfigure}{0.49\textwidth}
         \centering
         \includegraphics[width=\textwidth]{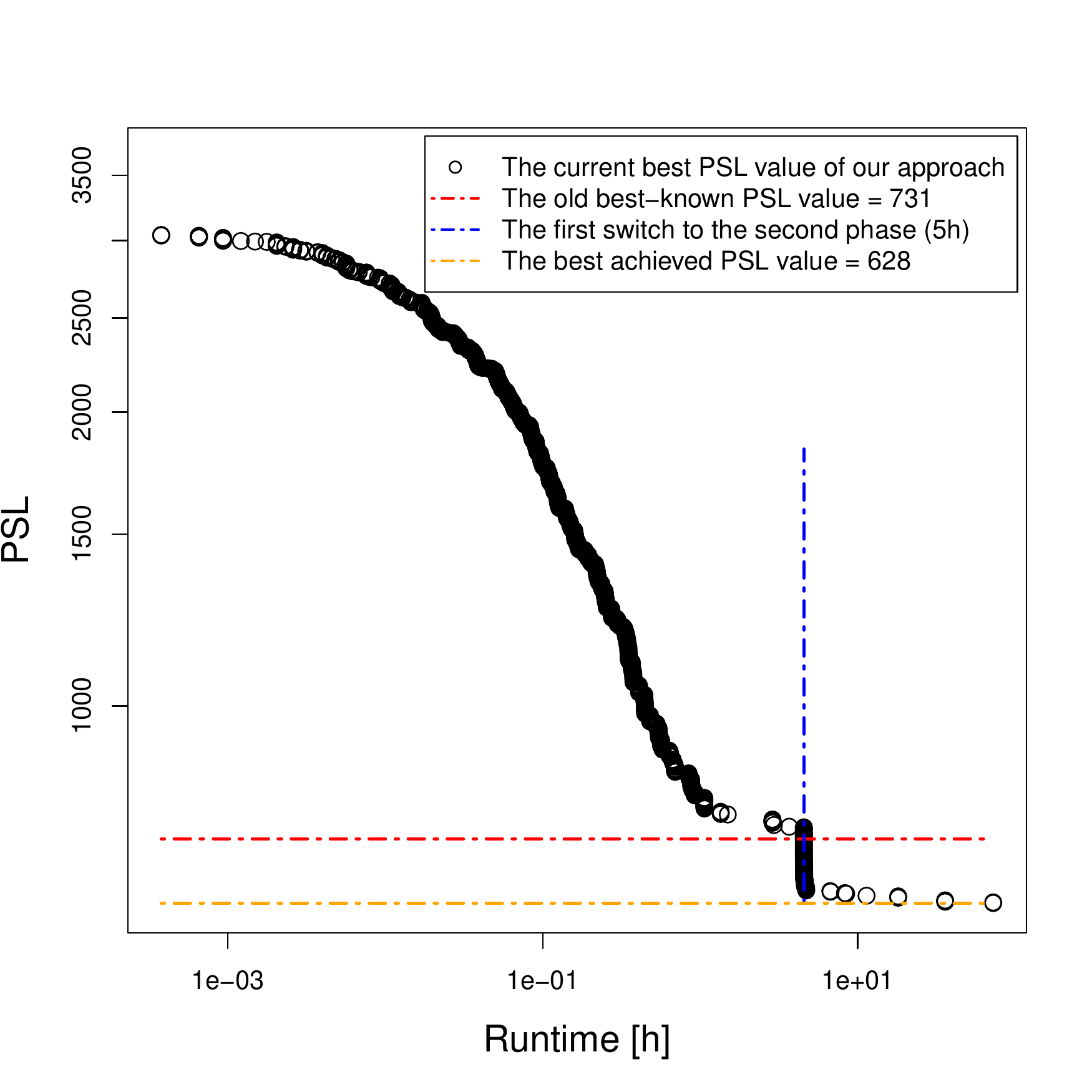}
         \caption{$L = 524287$}
         \label{fig:L524287}
     \end{subfigure}
     \hfill
     \begin{subfigure}{0.49\textwidth}
         \centering
         \includegraphics[width=\textwidth]{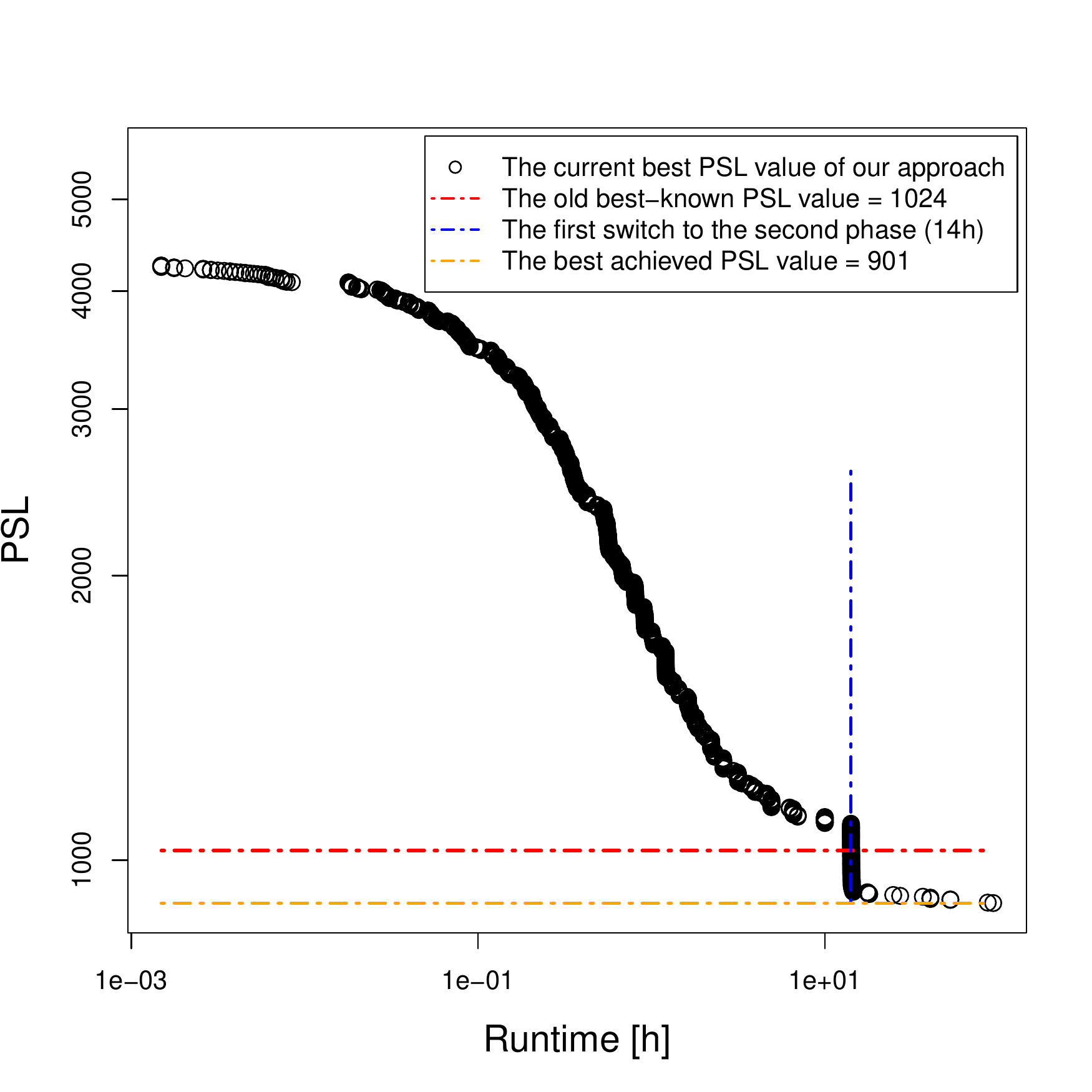}
         \caption{$L = 1048575$}
         \label{fig:L1048575}
     \end{subfigure}
     \caption{Convergence graphs.}
     \label{fig:con}
\end{figure*}

\begin{figure*}[t]
\centering
 \includegraphics[width=0.76\textwidth]{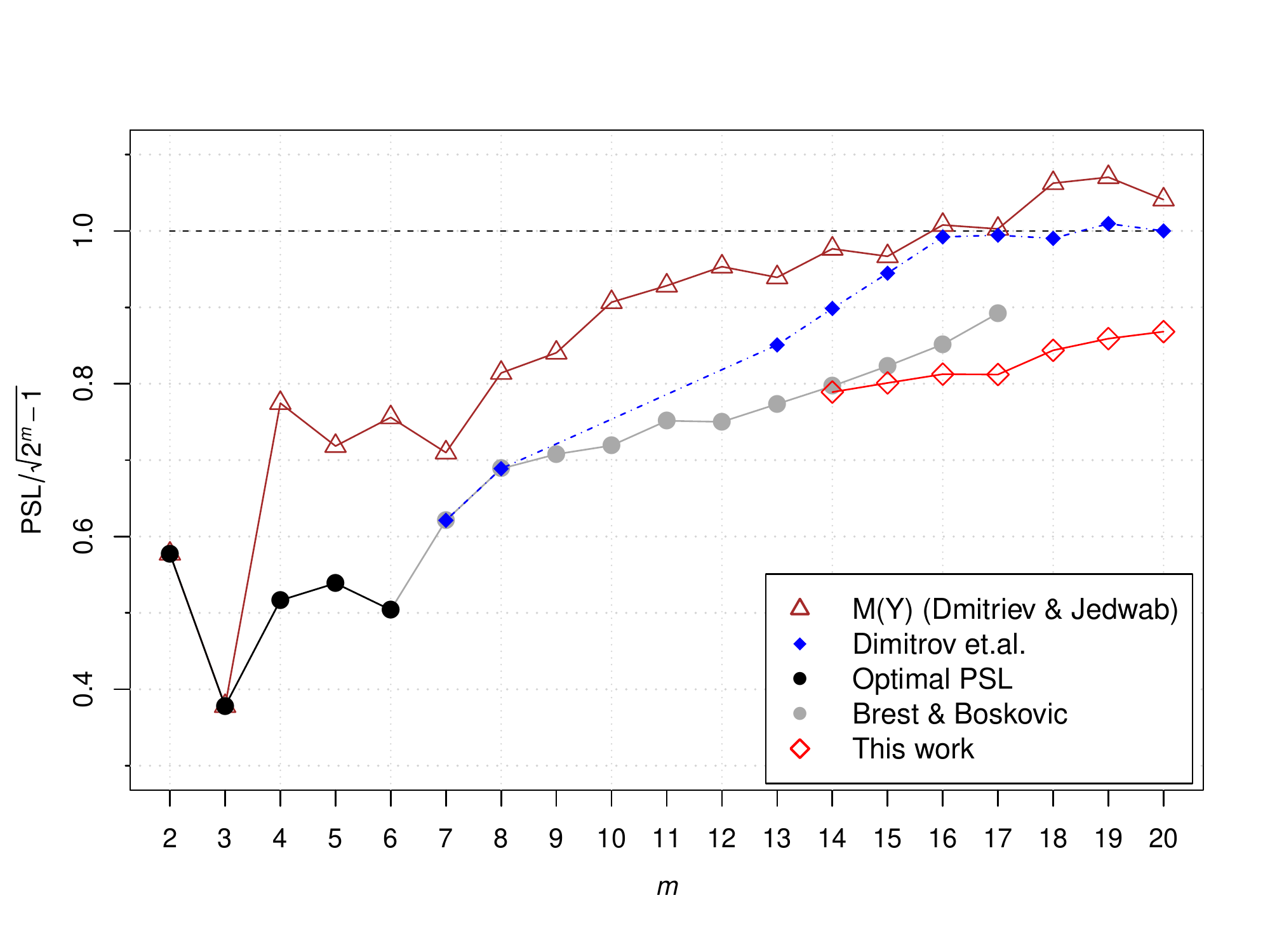}
 \caption{Comparison of the PSL value trends according to \cite{Dimitrov21a}, \cite{Brest21}, \cite{Dmitriev07}, and the best-known values.}
 \label{fig:lit}
\end{figure*}

The convergence graphs for the 4 largest instance sizes are shown in Fig.~\ref{fig:con} where the red line represents 
the old best-known PSL value, the orange line represents the best-achieved PSL values, and the blue line
represents the first switch to the second phase. We can observe that a significant improvement in PSL values
was achieved at the first switch to the second optimization phase, and these values were better in comparison with the old
best-known values. To achieve these values, the solver needed 0.4, 1, 5, and 14 hours for $L$ = 131071,
262143, 524287, and 1048575, respectively. In the continuation of the optimization process, these values improved further.
Therefore, we performed the next four runs for the all selected instance sizes. The same settings were used as before, but instead
to start the optimization process with a randomly seeded sequence the best-achieved sequence from the previous four runs was
selected as an initial sequence. The PSL values within these runs improved further, therefore, this process was repeated four times.
The achieved new best-known PSL values for all selected instance sizes are collected in Table~\ref{tab:nbk}. Here we can again observe that, as the instance
size increased, the difference between the old and new best-known PSL values ($\Delta $PSL) increased, and the
best improvement by 135 was achieved for the largest instance size. The old best-known results were obtained in~\cite{Brest21}
and~\cite{Dimitrov21a}. The first paper is our previous work, which was also based on a stochastic algorithm, while the exhaustive
search with restriction of $m$-sequences was used in the second paper. Additionally, a comparison of the growth rate of PSL
values, which also includes the results of~\cite{Dmitriev07}, is shown in Fig.~\ref{fig:lit}. This graph is based on instances that
belong to $m$-sequences, and it also includes the new best-known PSL values of our work. The PSL values achieved by the proposed algorithm were
significantly less than $\sqrt{L}$ (value 1.0 in the graph), significantly less in comparison with other works, the trend of these values is
now slightly flatter, and it is not possible to predict the growth rate for longer sequences.

\begin{table}[t]
\centering
\caption{Comparison of the best-known PSL values.}
\label{tab:nbk}
\resizebox{0.45\textwidth}{!}{
 \begin{tabular}{rr|rrr}
  $m$ &  \multicolumn{1}{c|}{$L$} & $\mathit{PSL}_{\mathit{old}}$ & $\mathit{PSL}_{\mathit{new}}$ & $\Delta \mathit{PSL}$\\
  \hline
  14 &   16383 &      102\,\cite{Brest21} & 101 & 1 \\
  15 &   32767 &      149\,\cite{Brest21} & 145 & 4 \\
  16 &   65535 &      218\,\cite{Brest21} & 208 & 10 \\
  17 &  131071 &      323\,\cite{Brest21} & 294 & 29 \\
  18 &  262143 &  507\,\cite{Dimitrov21a} & 432 & 75 \\
  19 &  524287 &  731\,\cite{Dimitrov21a} & 622 & 109 \\
  20 & 1048575 & 1024\,\cite{Dimitrov21a} & 889 & 135 \\
 \end{tabular}}
\end{table}

\section{Conclusion}
\label{sec:conclusion}
This paper introduces two-phase optimization of binary sequences with low peak sidelobe level value. The limited neighborhood search
is used in each optimization phase, with a corresponding fitness function that guides the search process. The fitness functions
differ in the value of the exponent, and the exponent determines which value of autocorrelation function has more influence on the final fitness
value. The optimization process switches between two optimization phases until the stopping condition is met. The peak sidelobe 
level (PSL) value is also calculated for all neighbors within a limited neighborhood search, and the sequence with the lowest
PSL value is returned at the end of the optimization process.

The proposed algorithm was implemented with the help of an efficient neighborhood structure and for concurrent computing on Graphics
Processing Units (GPU). This algorithm was tested on sequences with lengths $L = 2^m - 1$, for $14 \le m \le 20$. The experimental results 
show that the GPU solver achieved a speedup greater than 100. The two-phase optimization reduced the best-known
PSL values significantly, and as the instance size increased, this reduction was huge. For example, the PSL value was reduced
by 135 for $m = 20$.

In the future work, we will try to extend the proposed algorithm with more optimization phases, and within these phases we will try to
use different fitness functions, which will not only differ in exponents, but will be completely different.

\section*{Acknowledgements}
This work was supported by the Slovenian Research Agency (Computer Systems, Methodologies,
and Intelligent Services) under Grant P2-0041.

\bibliographystyle{spmpsci}


\begin{thebibliography}{10}
\providecommand{\url}[1]{{#1}}
\providecommand{\urlprefix}{URL }
\expandafter\ifx\csname urlstyle\endcsname\relax
  \providecommand{\doi}[1]{DOI~\discretionary{}{}{}#1}\else
  \providecommand{\doi}{DOI~\discretionary{}{}{}\begingroup
  \urlstyle{rm}\Url}\fi

\bibitem{Boskovic20}
Bošković, B., Brest, J.: Two-phase protein folding optimization on a
  three-dimensional ab off-lattice model.
\newblock Swarm and Evolutionary Computation \textbf{57}, 100,708 (2020).
\newblock \doi{https://doi.org/10.1016/j.swevo.2020.100708}

\bibitem{Boskovic17}
Bošković, B., Brglez, F., Brest, J.: Low-autocorrelation binary sequences: On
  improved merit factors and runtime predictions to achieve them.
\newblock Applied Soft Computing \textbf{56}, 262--285 (2017).
\newblock \doi{https://doi.org/10.1016/j.asoc.2017.02.024}

\bibitem{Brestarxiv}
Brest, J., Bo\v{s}kovi\'{c}, B.: In searching of long skew-symmetric binary
  sequences with high merit factors.
\newblock CoRR \textbf{abs/2011.00068} (2020).
\newblock \urlprefix\url{https://arxiv.org/abs/2011.00068}

\bibitem{Brest21}
Brest, J., Bošković, B.: Low autocorrelation binary sequences: Best-known
  peak sidelobe level values.
\newblock IEEE Access \textbf{9}, 67,713--67,723 (2021).
\newblock \doi{10.1109/ACCESS.2021.3077541}

\bibitem{Coxson05}
Coxson, G., Russo, J.: Efficient exhaustive search for optimal-peak-sidelobe
  binary codes.
\newblock IEEE Transactions on Aerospace and Electronic Systems \textbf{41}(1),
  302--308 (2005).
\newblock \doi{10.1109/TAES.2005.1413763}

\bibitem{Longqing21}
Cui, L., Liu, X., Lu, S., Jia, Z.: A variable neighborhood search approach for
  the resource-constrained multi-project collaborative scheduling problem.
\newblock Applied Soft Computing \textbf{107}, 107,480 (2021).
\newblock \doi{j.asoc.2021.107480}

\bibitem{Dimitrov21a}
Dimitrov, M.: On the aperiodic autocorrelations of rotated binary sequences.
\newblock IEEE Communications Letters \textbf{25} (2021).
\newblock \doi{10.1109/LCOMM.2020.3047899}

\bibitem{Dimitrov20a}
Dimitrov, M., Baitcheva, T., Nikolov, N.: Efficient generation of low
  autocorrelation binary sequences.
\newblock IEEE Signal Processing Letters \textbf{27}, 341--345 (2020).
\newblock \doi{10.1109/LSP.2020.2972127}

\bibitem{Dimitrov20b}
Dimitrov, M., Baitcheva, T., Nikolov, N.: On the generation of long binary
  sequences with record-breaking psl values.
\newblock IEEE Signal Processing Letters \textbf{27}, 1904--1908 (2020).
\newblock \doi{10.1109/LSP.2020.3031463}

\newpage

\bibitem{Dimitrov21b}
Dimitrov, M., Baitcheva, T., Nikolov, N.: Hybrid constructions of binary
  sequences with low autocorrelation sidelobes.
\newblock arXiv preprint arXiv:2104.10477  (2021)

\bibitem{Dmitriev07}
Dmitriev, D., Jedwab, J.: Bounds on the growth rate of the peak sidelobe level
  of binary sequences.
\newblock Adv. Math. Commun. \textbf{1}(4), 461--475 (2007).
\newblock \doi{10.3934/amc.2007.1.461}

\bibitem{Gallardo09}
Gallardo, J.E., Cotta, C., Fernández, A.J.: Finding low autocorrelation binary
  sequences with memetic algorithms.
\newblock Applied Soft Computing \textbf{9}(4), 1252--1262 (2009).
\newblock \doi{https://doi.org/10.1016/j.asoc.2009.03.005}

\bibitem{golomb05}
Golomb, S.W., Gong, G.: Signal design for good correlation: for wireless
  communication, cryptography, and radar.
\newblock Cambridge University Press (2005)

\bibitem{kroszczynski69}
Kroszczynski, J.: Pulse compression by means of linear-period modulation.
\newblock Proceedings of the IEEE \textbf{57}(7), 1260--1266 (1969).
\newblock \doi{10.1109/PROC.1969.7230}

\bibitem{Leukhin13}
Leukhin, A.N., Potekhin, E.N.: Optimal peak sidelobe level sequences up to
  length 74.
\newblock In: 2013 European Microwave Conference, pp. 1807--1810 (2013).
\newblock \doi{10.23919/EuMC.2013.6687030}

\bibitem{Song15}
Song, J., Babu, P., Palomar, D.P.: Optimization methods for designing sequences
  with low autocorrelation sidelobes.
\newblock IEEE Transactions on Signal Processing \textbf{63}(15), 3998--4009
  (2015).
\newblock \doi{10.1109/TSP.2015.2425808}

\bibitem{Vincent21}
Yu, V.F., Jodiawan, P., Gunawan, A.: An adaptive large neighborhood search for
  the green mixed fleet vehicle routing problem with realistic energy
  consumption and partial recharges.
\newblock Applied Soft Computing \textbf{105}, 107,251 (2021).
\newblock \doi{10.1016/j.asoc.2021.107251}

\bibitem{Biao21}
Zhang, B., Pan, Q.K., Meng, L.L., Zhang, X.L., Ren, Y.P., Li, J.Q., Jiang,
  X.C.: A collaborative variable neighborhood descent algorithm for the hybrid
  flowshop scheduling problem with consistent sublots.
\newblock Applied Soft Computing \textbf{106}, 107,305 (2021).
\newblock \doi{10.1016/j.asoc.2021.107305}

\bibitem{Dominik21}
\.{Z}urek, D., Pi\k{e}tak, K., Pietro\'{n}, M., Kisiel-Dorohinicki, M.: New
  variants of sdls algorithm for labs problem dedicated to gpgpu architectures.
\newblock In: M.~Paszynski, D.~Kranzlm{\"u}ller, V.V. Krzhizhanovskaya, J.J.
  Dongarra, P.M.A. Sloot (eds.) Computational Science -- ICCS 2021, pp.
  206--212. Springer International Publishing, Cham (2021)

\end{thebibliography}

\end{document}